\documentclass[journal]{IEEEtran}
\ifCLASSINFOpdf
\else
\fi
\usepackage{mathtools}
\usepackage{color}
\usepackage{algorithm}
\usepackage[noend]{algpseudocode}

\usepackage{times}
\usepackage{epsfig}
\usepackage{graphicx}
\usepackage{amssymb}
\usepackage{romannum}
\usepackage{color}

\usepackage{amsmath}
\usepackage{breqn}

\begin{document}
\title{Integrating Human-in-the-loop into Swarm Learning for Decentralized Fake News Detection}

 \author{Xishuang Dong, and~Lijun Qian,~\IEEEmembership{Senior Member,~IEEE,}
        % <-this % stops a space
\thanks{X. Dong and L. Qian are with the Center of Excellence in Research and Education for Big Military Data Intelligence (CREDIT Center), Department of Electrical and Computer Engineering, Prairie View A\&M University, Texas A\&M University System, Prairie View, TX 77446, USA. Email: xidong@pvamu.edu, liqian@pvamu.edu}% <-this % stops a space
}

\maketitle

\begin{abstract}
Social media has become an effective platform to generate and spread fake news that can mislead people and even distort public opinion. Centralized methods for fake news detection, however, cannot effectively protect user privacy during the process of centralized data collection for training models. Moreover, it cannot fully involve user feedback in the loop of learning detection models for further enhancing fake news detection.  To overcome these challenges, this paper proposed a novel decentralized method, Human-in-the-loop Based Swarm Learning  (HBSL),  to integrate user feedback into the loop of learning and inference for recognizing fake news without violating user privacy in a decentralized manner. It consists of distributed nodes that are able to independently learn and detect fake news on local data. Furthermore, detection models trained on these nodes can be enhanced through decentralized model merging. Experimental results demonstrate that the proposed method outperforms the state-of-the-art decentralized method in regard of detecting fake news on a benchmark dataset.  
 \end{abstract}


% \fi
\IEEEpeerreviewmaketitle

\section{Introduction}
\label{sec1}
% introduction 

 \begin{figure*} [ht]
 	\centering
	\includegraphics[width=.9\linewidth]{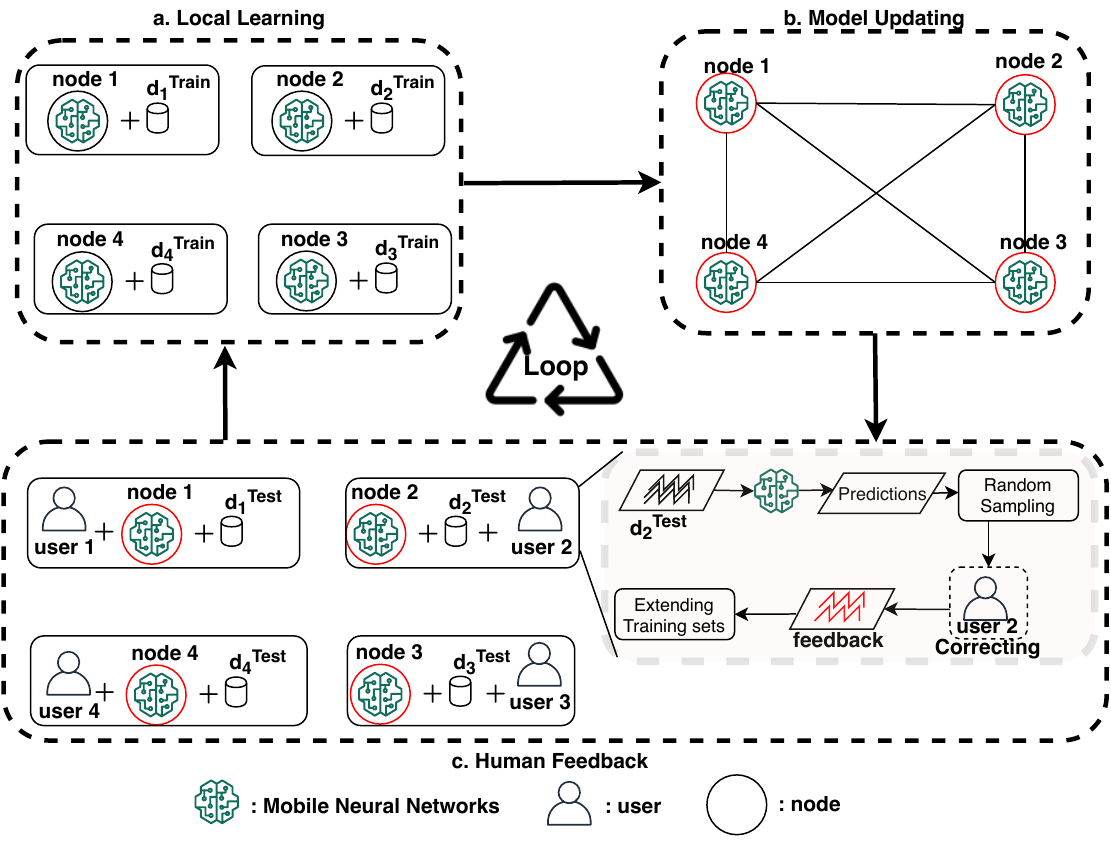}
	\caption{Diagram of human-in-the-loop based swarm learning (HBSL) for fake news detection. It is composed of three stages, namely, (a) Local Learning, (b) Model Updating, and (c) Human Feedback, where the training set $d_{i}^{Train}$ for node $i$ will be different from $d_{j}^{Train}$ for node $j$, and the testing sets $d_{i}^{Test}$  and $d_{j}^{Test}$ for nodes $i$ and $j$, respectively, are different as well, $ 1 \leqslant i \leqslant 4$, and $ 1 \leqslant j \leqslant 4$. In stage (a), all nodes will learn the detection models on the local data. Then, in stage (b), these models will be updated by a master node selected through averaging model weights. Finally, in stage (c), all nodes will apply the model updated to accomplish fake news detection on the local testing data. Afterwards, users will correct a portion of predictions selected by random sampling as feedback to extend the training data. The models updated will be enhanced by fine-tuning on the training sets extended. All these three stages will form a loop of learning and inference to update their models until meeting the stop criteria. }
	\label{Fig1_framework}
\end{figure*}

The development of social media (e.g., Twitter) has significantly changed the way of information collection for human. Social media has dominated information generation and spreading, which results in that it is becoming more and more challenging for people to live without social media~\cite{imran2015processing, pennycook2019fighting, phuvipadawat2010breaking}. Although it reduces the cost of information retrieval, unfortunately, the absence of systematic and effective management on the information on social media platforms has led to that social media comes up with the hotbed of generation and spreading of fake news~\cite{webb2016digital, wu2019misinformation}, where fake news refers to the news that is intentionally and verifiably false~\cite{allcott2017social, zhou2020survey}. As a result, fake news causes many confusions and severe damages to the society. For example, during the COVID-19 pandemic, fake news makes it more difficult for people to find trustworthy and reliable information to combat the spreading and treatment of the virus~\cite{geleris2020observational}. Thus, preventing the spreading of fake news is imperative to decrease political polarization, increase trust in public institutions, and improve decision-making for everyone’s life.

Machine learning techniques are effective to recognize potential fake news by building models on news features including content~\cite{conroy2015automatic, rubin2016fake, ruchansky2017csi} and context~\cite{tacchini2017some, papanastasiou2020fake, shu2020hierarchical}. Centralized detection methods dominated this field by collecting big data to a cloud storage for building high-performance detection models, where deep learning techniques such as convolutional neural networks (CNN), recurrent neural networks (RNN), and deep graph models outperform other techniques~\cite{ruchansky2017csi, ajao2018fake, song2021temporally, ren2020adversarial, monti2019fake, benamira2019semi}. Unfortunately, one potential risk in the procedure of building these methods is to violate user privacy when collecting big data of news in the centralized manner. 

To reduce this risk, decentralized methods are required to implement privacy preserving, which is to learn and infer on local data, not upload data to a centralized data storage for building detection models. Federated learning~\cite{konevcny2016federated, yang2019federated, li2020federated} is  a distributed machine learning approach that enables training a high-quality centralized model while training data remains distributed over a large number of users. However, it relies on a model center to control the process of updating models for users, which increases the potentials of hacking regarding the communication between the model center and user models. On the contrary, swarm learning~\cite{warnat2021swarm} is able to implement decentralized learning to maintain user privacy without the need for a central coordinator, thereby going beyond federated learning. It enables decentralized training without sharing the data through learning a set of nodes, where each node learns on training data locally and enhances the model collaboratively without sharing the training data. They share parameters (weights) derived from training the model on the local data. Thus, it allows users at the nodes to maintain the confidentiality and privacy of the raw data. Nevertheless, swarm learning is not designed to leverage valuable user feedback on the inference to further enhance learning performance since the user feedback has been approved to effective to improve data analysis in the loop of learning and inference~\cite{wu2021survey}. 

In this paper, we propose a human-in-the-loop based based swarm learning (HBSL) to integrate user feedback into the learning and inference of swarm learning via human-in-the-loop (HITL) techniques~\cite{wu2021survey}. HITL is to involve human activities in the process of building machine learning models to improve the model performance via human knowledge~\cite{wu2021survey}. We applied HITL to generate the user feedback in the learning process to improve fake news detection. The detailed learning process is shown in the diagram shown in Figure~\ref{Fig1_framework}. It consists of three stages in the learning process, namely, (a) Local Learning, (b) Model Updating, and (c) Human Feedback, which forms a loop until the learning is terminated when meeting stop criteria. In stage (a), all nodes will learn the detection models independently on the local data. Then these models in all nodes will be updated in the stage (b) by a master node through averaging model weights. In stage (c), all nodes will apply the model updated to accomplish fake news detection on the local testing data. Afterwards, users will provide feedback on the predictions of testing data, which is used to extend the training data to improve the detection. All models on these nodes will keep updating their models in a loop with these three stages until meeting the stop criteria of learning. Experimental results demonstrate that the proposed method outperformed swarm learning on decentralized fake news detection.

In summary, the contributions of this study are:
\begin{itemize}
\item We proposed a novel decentralized method through combing swarm learning and human-in-the-loop. In the learning and inference process, users in different nodes provides feedbacks on the inference results to generate feedback. Then nodes collects the feedback to extend training sets to fine-tune the model in order to enhance the inference performance within a loop of learning and inference.
\item We validate our proposed model on a benchmark LIAR~\cite{wang2017liar}. Compared to swarm learning, the proposed model is able to significantly improve the performance for each node on detecting fake news by learning on local training data together with user feedbacks. 
\end{itemize}

\section{Model}
\label{sec4}
%method

The proposed decentralize method is to combine the advantages of swarm learning~\cite{warnat2021swarm}  and human-in the loop (HITL)~\cite{wu2021survey}, which is able to accomplish decentralized fake news detection via human feedback and model update in the swarm learning.

\subsection{Swarm Learning}

\begin{figure} [ht]
	\centering
	\includegraphics[scale=0.45]{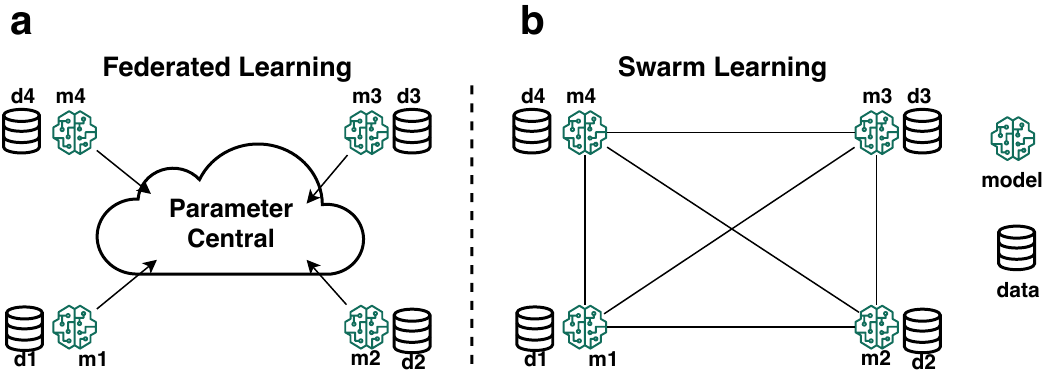}
	\caption{Comparison between swarm learning and federated learning. $d_{i}$ is different from $d_{j}$, where $ 1 \leqslant i \leqslant 4$, and $ 1 \leqslant j \leqslant 4$. Federated learning keeps data with local data contributors and perform learning at the site of local data storage and availability, but control parameter updating by a central parameter server.  On the contrary, swarm learning doesn’t need parameter central control for parameter update  since the data and parameters of a model will be stored locally. }
	\label{Fig_swarm_learning}
\end{figure}

Swarm learning~\cite{warnat2021swarm} is a decentralized machine-learning approach that unites edge computing, blockchain-based peer-to-peer networking and coordination while maintaining privacy without a central control. It implements learning on distributed nodes (edges) without sharing data to protect privacy in a local community.  Compared to federated learning~\cite{konevcny2016federated, konevcny2016federated2}, swarm learning will be totally decentralized without parameter central control, which is illustrated as Figure~\ref{Fig_swarm_learning}. 

In addition, swarm learning can update parameters without complex process to enhance the inference performance on nodes, only through merging parameters of models between nodes given by

\begin{equation}
	P_{M} = \frac{\sum_{k=1}^{n}(w_{k} \times P_{k})}{n \times \sum_{k=1}^{n}w_{k}}.
\end{equation}

where $P_{M}$ is  the model parameter updated on a node, $P_{k}$  is parameters from the $k^{th}$ node, $w_{k}$ is the weight of the $k^{th}$ node, and $n$ is the number of nodes participating in the merging process.

\subsection{Human-in-the-Loop (HITL)}

\begin{figure} [ht]
	\centering
	\includegraphics[scale=0.45]{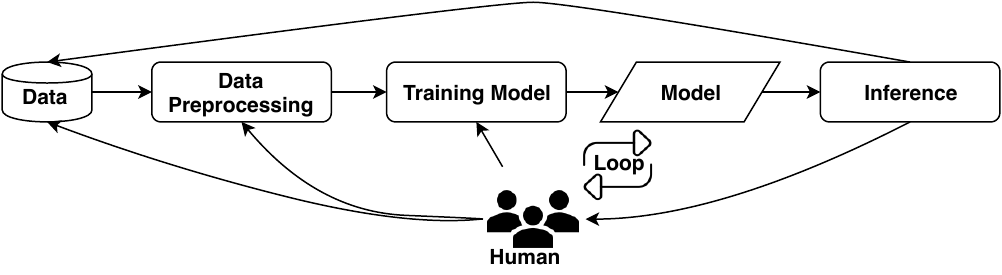}
	\caption{Framework of human-in-the-loop for machine learning. }
	\label{Fig_hitl}
\end{figure}

Human-in-the-loop can be applied to improve the performance of machine learning models by integrating human knowledge and experience for data analytics~\cite{wu2021survey}. For example, human can significantly reduce algorithm bias in the training and inference in terms of human feedback  for various tasks in the field of natural language processing (NLP) such as text classification~\cite{karmakharm2019journalist}, syntactic and semantic parsing~\cite{su2017building}, topic modeling~\cite{hu2014interactive}, text summarization~\cite{stiennon2020learning}, and sentiment analysis~\cite{liu2021and}. The general framework is shown in Figure~\ref{Fig_hitl}.

Human can provide feedback to the model training, data preprocessing, data collection regarding the model inference, even predictions to improve the model inference in a loop, where the feedback can be associate with inference results and its performance, inference time cost, and inference computation cost. 

\subsection{HITL Based Swarm Learning (HBSL)}

This paper proposed a model to combine swarm learning and HITL to implement decentralized fake news detection. Figure~\ref{Fig1_framework} presents the flow of building the proposed model with four nodes in three stages in the learning process, namely, (a) Local Learning, (b) Model Updating, and (c) Human Feedback, which forms a loop until the learning is terminated.

\textit{Local Learning}:  This stage is to independently learn the model on local data in the node. The local data is preprocessed by removing missing values, stemming, and one-hot representations of sentences. These representations are input to the model built via bidirectional recurrent neural networks (BRNN). This model will not be complex regarding the resource constrain like limited memory and thus contains one embedding layer, one forward layer, and one backward layers. The embedding layer represents the input  $<$$x_{1}, x_{2}, x_{3}, ..., x_{t}, ..., x_{n}$$>$ as an embedding vector $<$$e_{1}, e_{2}, e_{3}, ..., e_{t}, ..., e_{n}$$>$, which is to resolve feature sparsity issues. Afterwards, these two layers generate two directional correlation features on the embedding vector. Then we combine these two features as the output $z$  of this model, where $z$ is a sequence $<$$z_{1}, z_{2}, z_{3}, ..., z_{t}, ..., z_{n}$$>$ and $z_{t}$ is given by

\begin{dmath}
	z_{t} =  w_{z}^{f}h_{t}^{f} + w_{z}^{b}h_{t}^{b} + b_{z} \; ,
\label{Equ_output_shared_rnn}
\end{dmath}

where 

\begin{dmath}
	h_{t}^{f} =  a(w_{h}^{f}h_{t - 1}^{f} + w_{e}^{f}e_{t} + b_{h}^{f}) \; ,
\label{Equ_output_h_rnn}
\end{dmath}

\begin{dmath}
	h_{t}^{b} =  a(w_{h}^{b}h_{t + 1}^{b} + w_{e}^{b}e_{t} + b_{h}^{b}) \; ,
\label{Equ_output_h_rnn}
\end{dmath}

$z_{t}$ is the output of $x_{t}$ for the input $<$$x_{1}, x_{2}, x_{3}, ..., x_{t}, ..., x_{n}$$>$.  $a(\cdot)$ is the activation function for hidden layers. $w_{z}^{f}$, $w_{h}^{f}$, and $w_{e}^{f}$ are forward weights for three layers, namely, output layer, forward layer, and backward layer. $w_{z}^{b}$, $w_{h}^{b}$, and $w_{e}^{b}$ are backward weights for these three layers, respectively. $b_{z}$, $b_{h}^{f}$, and $b_{h}^{b}$ are bias for these three layers. 

We utilize the output $z$ to calculate the cross entropy loss given by 

\begin{dmath}
	Loss =  -y \times log \phi{(z)} - (1- y) \times log (1- \phi{(z)}) \; ,
\label{Equ_loss}
\end{dmath}

where $\phi{(z)}$ is the sigmod function. These four nodes will learn these models with this identical architecture of BRNN on local data.

\textit{Model Updating}: The master node in this stage will be selected in terms of predefined criterion. The criterion is that the node with the highest performance of detection is selected as the master node, which can be implemented by local communication between nodes. This node collects all models from other nodes and updates parameters $w_{master}$ below.

\begin{equation}
	w'_{master} = \frac{\sum_{i =1}^{n}w_{i} }{n}.
\label{update_weight}	
\end{equation}

where $n = 4$ in this proposed method. Afterwards, the master node will share the updated model to replace parameters of models in other nodes.

\textit{Human Feedback}: In the third stage, the model will perform inference on their local testing data in a parallel manner, where the dataset for this inference for different nodes will be various. For example, for the diagram in Figure~\ref{Fig1_framework}, the dataset $d_{i}^{Test}$ for node $i$ will be different from $d_{j}^{Test}$ for node $j$, where $i \neq j$, $ 1 \leqslant i \leqslant 4$, and $ 1 \leqslant j \leqslant 4$. Then, each user $i$ (Human) will generate feedback on the prediction results. In detailed, the user will randomly select a portion of predictions and correct them as feedback. This feedback will be integrated into corresponding training data to extend training sets. For instance, feedback on the predictions of $d_{2}^{Test}$ will generate a set of corrected predictions $d'_{2}$. Then $d'_{2}$ will be integrated into $d_{2}^{Train}$ to extend the training set for next round of local learning of local learning in the loop of learning.

Stages (a) to (c) form the loop to update node's model parameters to enhance detection performance. The details learning process is shown in Algorithm~\ref{Arg1_learning}, where the portion of predictions is predefined.

\begin{algorithm}[ht]
	\caption{Learning in the proposed method}
	\begin{algorithmic}[1]
		\For{$t$ in~[1, rounds] }
				\State{Learning on each node's data $d_{t, i}^{Train}$, $ 1 \leqslant i \leqslant 4$}
				\State{Updating the parameter $w_{t, i}$ of the model of the master node $i$ with equation~(\ref{update_weight}) }
				\State{Replacing the model parameter of other nodes with $w_{t, i}$}
			 	\State{Inferencing on each node's data $d_{t, i}^{Test}$, $ 1 \leqslant i \leqslant 4$}
				\State{Generating feedback $d'_{t, i}$ by correcting a portion of predictions  on each nodes, $ 1 \leqslant i \leqslant 4$}
				\State{Extending $d_{t, i}^{Train}$  = \{$d_{t, i}^{Train}$, $d'_{t, i}$\}, $ 1 \leqslant i \leqslant 4$}			 
		\EndFor
	\Return{$w_{i}$ for node $i$, $ 1 \leqslant i \leqslant 4$}
	\end{algorithmic}
	\label{Arg1_learning}
\end{algorithm}

\section{Experiment}
\label{sec5}
% experiments

\subsection{Datasets}

We validate the effectiveness of the proposed model by detecting fake news on the benchmark LIAR~\cite{wang2017liar}. It is a standard benchmark dataset for fake news detection. It includes 12,836 real-world short statements collected from a variety of occasions such as debate, campaign, Facebook, Twitter, interviews, ads, etc. Each statement is labeled with six-grade truthfulness, namely, true, false, half-true, part-fire, barely-true, and mostly-true. We reorganize the data as two classes by treating five classes including false, half-true, part-fire, barely-true, and mostly-true as \textbf{Fake} class and true as \textbf{True} class. Therefore, the fake news detection on this benchmark is converted to a binary classification task.

\subsection{Experiment Setup}

The key hyper-parameters for training the proposed model are shown in Table~\ref{tab_hyperpara} and we employ Adam optimizer to complete the training.
\begin{table} [ht]
	\caption{\label{tab_hyperpara} Hyper-parameters for the learning of the model on nodes}
        \begin{center}
                \begin{tabular}{|l|r|}
                 \hline \textbf{Hyper-parameters} & \textbf{Values} \\ \hline
                	Embedding size		& 40 \\
		Batch size				& 32 \\
		Number of epochs		& 10 \\
		Learning rate 			& 1e-3 \\
		Feedback portion  		& 0.2 \\
		\hline
               \end{tabular}
       \end{center}
 \end{table}      

\subsection{Evaluation}
We apply accuracy to evaluate the performance of fake news detection regarding the task features on the benchmark, where the accuracy is calculated by dividing the number of news detected correctly over the total number of news. 

\begin{equation}
	Accuracy = \frac{N_{correct}}{N_{total}}.
\end{equation}

\subsection{Results}

We validate the proposed methods with two different configurations. One is to implement decentralized learning on four nodes with human feedback while the other is to implement learning on eight nodes.

%% %%%%%%%%%%%%%% 4 Node Experiment Results %%%%%%%%%%%%%%%%%
%%%%%%%%%%%%%%%%%%%%%%%%%%%%%%%%%%%%%%%%%%%%%%%

\subsubsection{Learning on four nodes} 
\begin{figure} [ht]
 	\centering
	\includegraphics[width=.9\linewidth]{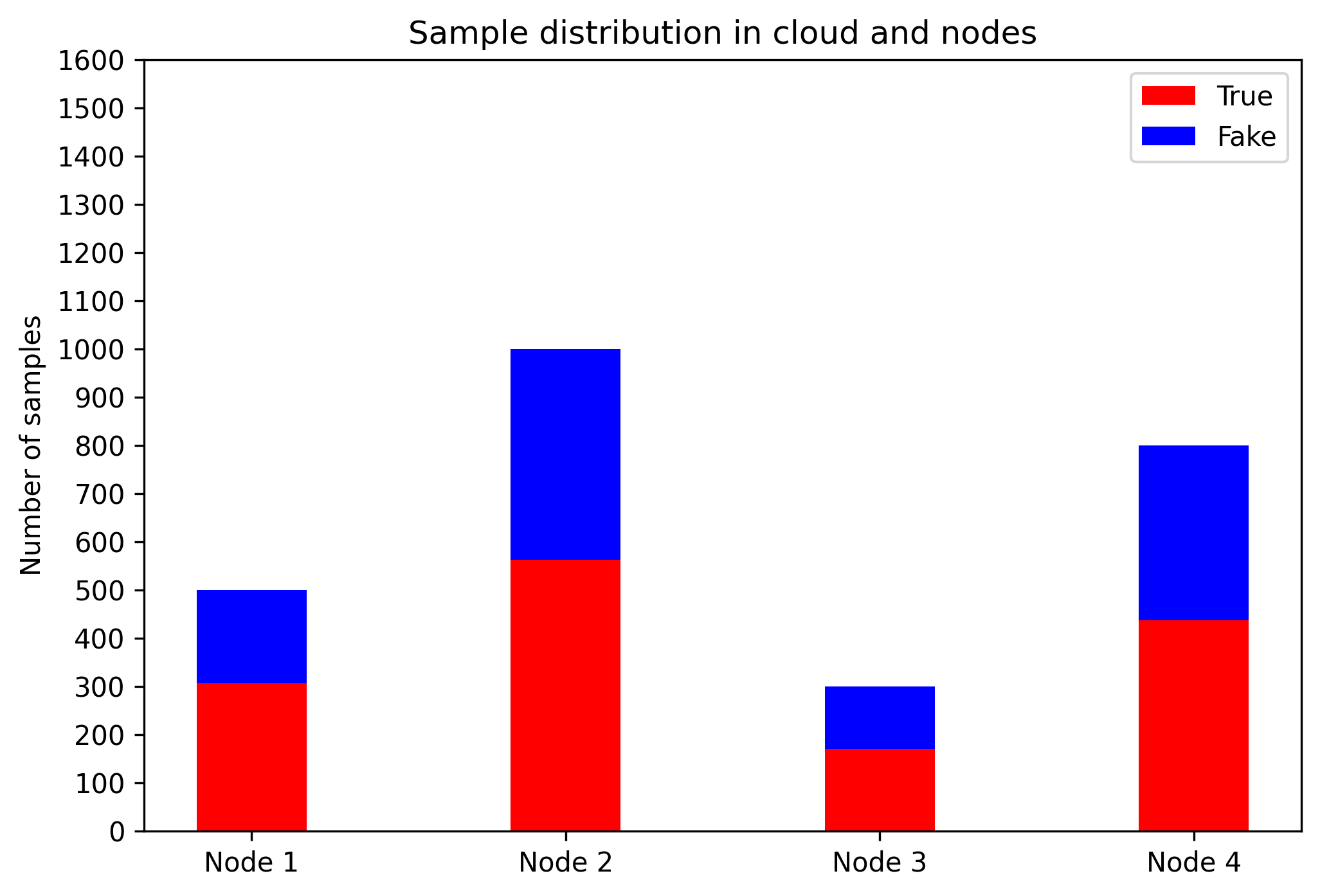}
	\caption{Sample distribution on 4 nodes for fake news detection. }
	\label{Fig_4_nodes_data}
\end{figure}

In that regard of limited data available for individual user in the real application, we distribute small number of samples (e.g. less than $1,000$) for inference in different nodes, which is shown in Figure~\ref{Fig_4_nodes_data} with two classes: Fake and True. Specifically, we assume that all nodes share the same class distribution to simplify the task.

\begin{table}[ht]
	\caption{ Accuracy (\%) comparison on fake news detection for learning with 4 nodes. } 
        \begin{center}
                \begin{tabular}{|l|ccccc|}
                    \hline \textbf{SL} & Node 1 & Node 2 & Node 3 & Node 4 & \textbf{Average}\\ \hline
                    	   Run 1		&  86.40\%	&  86.40\%	&  90.66\%	&  91.50\%	&  88.74\% \\
                            Run 2		&  89.60\%	&  89.60\%	&  93.33\%	&  85.50\%	&  89.50\%\\
                            Run 3		&  87.99\%	&  89.60\%	&  86.66\%	&  86.00\%	&  87.56\%\\
                            Run 4		&  82.40\%	&  85.60\%	&  85.33\%	&  85.50\%	&  84.70\%\\
                            Run 5		&  86.40\%	&  89.60\%	&  93.33\%	&  85.50\%	&  88.70\%\\ \hline
                            \textbf{Average}		&  86.55\%	&  88.16\%	&  89.86\%	&  86.80\%	&  87.84\%\\  
                     \hline
                     \hline \textbf{HBSL} & Node 1 & Node 2 & Node 3 & Node 4 & \textbf{Average}\\ \hline
                    	   Run 1		&  92.00\%	&  89.99\%	&  92.00\%	&  94.99\%	&  92.24\%\\
                            Run 2		&  90.39\%	&  92.79\%	&  97.33\%	&  90.49\%	&  92.75\%\\
                            Run 3		&  92.79\%	&  94.40\%	&  87.99\%	&  91.00\%	&  91.55\%\\
                            Run 4		&  85.60\%	&  89.20\%	&  87.99\%	&  89.99\%	&  88.19\%\\
                            Run 5		&  90.39\%	&  89.20\%	&  94.66\%	&  91.00\%	&  91.31\%\\ \hline
                            \textbf{Average}		&  90.23\%	&  91.11\%	&  91.99\%	&  91.49\%	&  91.20\%\\ 

                     \hline
                \end{tabular}
       \end{center}
         \label{tab_detail_4nodes}
\end{table}

We show the details of performance of 5 runs in Table~\ref{tab_detail_4nodes} from two dimensions. One dimension is to examine the performance in terms of the accuracy for different runs. It appears that HBSL consistently performs better than SL in different runs. The other dimension is to check the performance for different nodes. HBSL can improve detection performance for different nodes up to 5\%. From these two dimensions, human feedback for swarm learning can consistently improve detection performance for 5 runs of 4 nodes. However, it can be observed that the performance of these nodes for different runs are various up to 7\% and 10\% for SL and HBSL, respectively. The unstable performance is caused by learning on small local data with the shallow RNNs on these nodes. 

\begin{figure} [ht]
	\centering
	\includegraphics[scale=0.22]{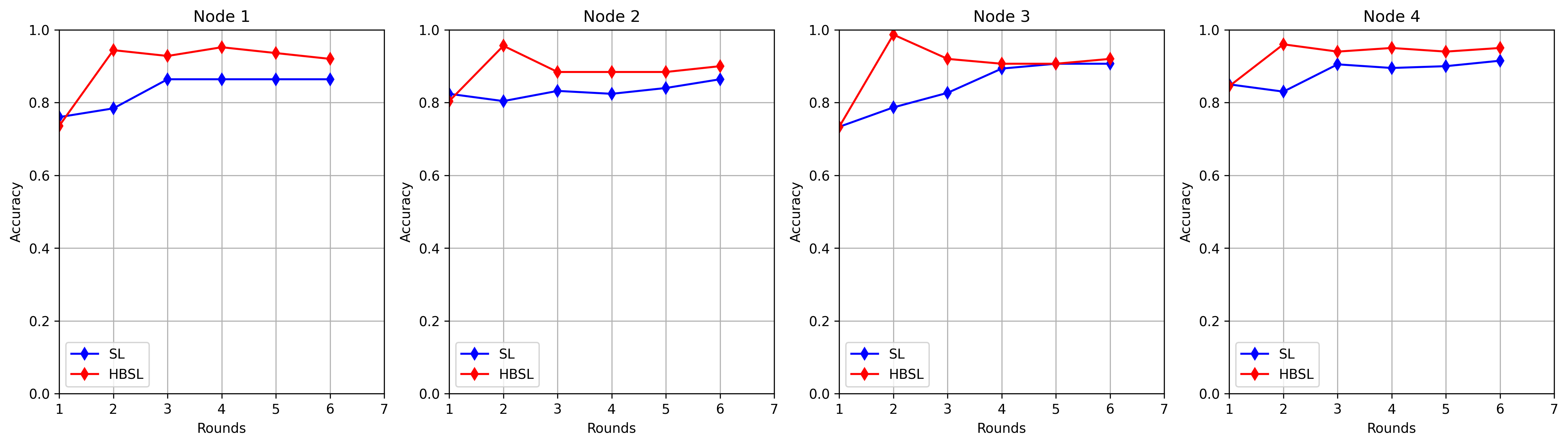} \\(a) Run 1\\
	\includegraphics[scale=0.22]{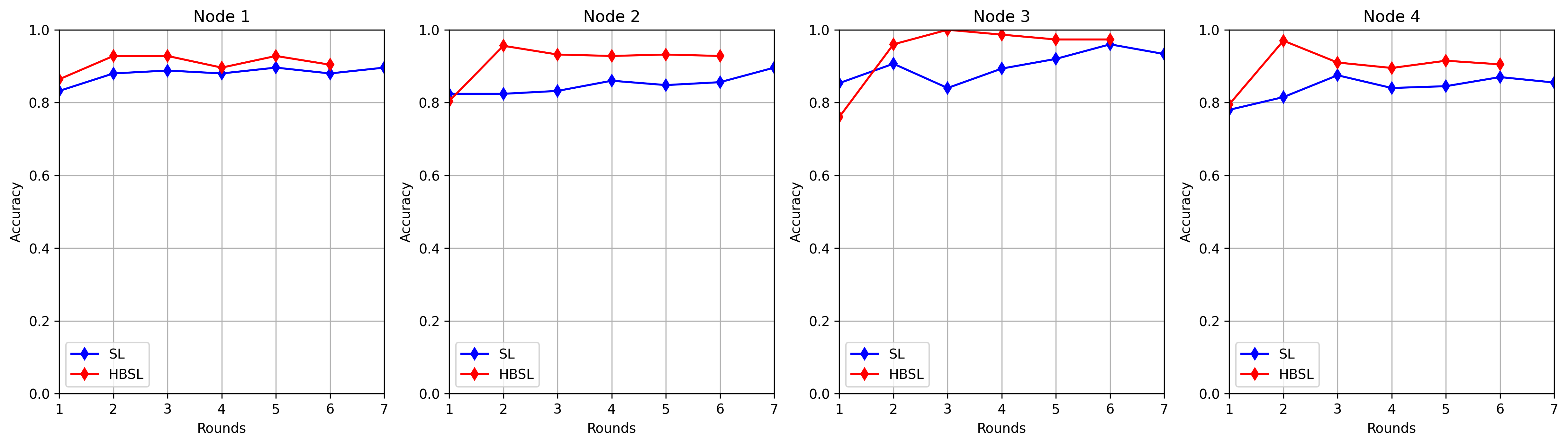} \\(b) Run 2 \\
	\includegraphics[scale=0.22]{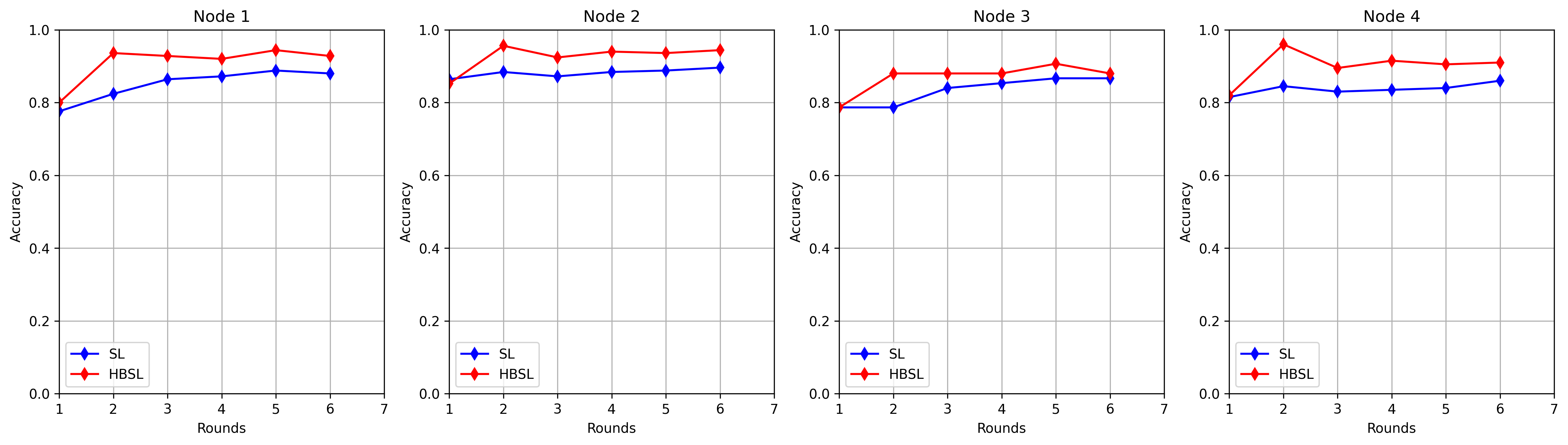} \\(c) Run 3\\ 
	\includegraphics[scale=0.22]{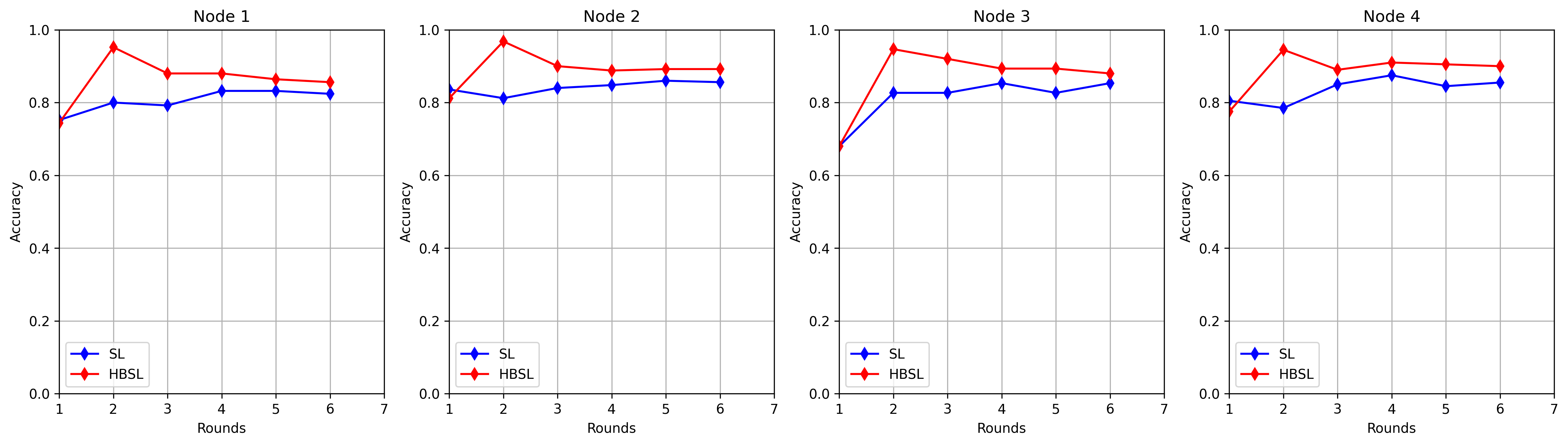} \\(d) Run 4\\
	\includegraphics[scale= 0.22]{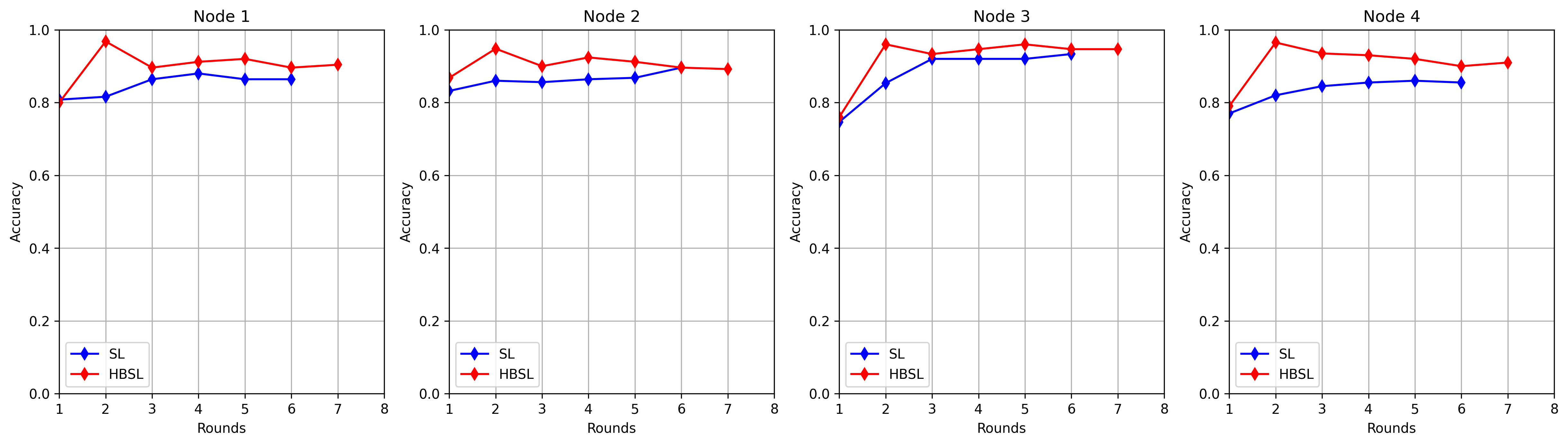} \\(e) Run 5\\
	\caption{Inference performance for 4 nodes in the learning process for HBSL in 5 runs. Accuracy refers to testing accuracy.}
	\label{Fig_4nodes_accuracy}
\end{figure}

Moreover, Figure~\ref{Fig_4nodes_accuracy}  presents the inference performance for different nodes in the learning process. It is observed that with human feedbacks, HBSL detection performance is higher than those of SL, which means that human feedback can contribute to improvement of detection performance.  In addition, compared to SL, HBSL can converge to higher performance for most of nodes in different runs. Moreover, although human feedback enlarges the training sets by introducing more samples, it will not decrease learning speed by increasing learning rounds. 

%\begin{figure*} [ht]
%	\centering
%	\includegraphics[scale=0.45]{Figure/results/4nodes/node_accuracy_1.png}
%	\caption{Testing accuracy for different nodes in the learning process for SLHITL. Rounds refer to the learning iterations in Algorithm 1.}
%	\label{Fig_4nodes_accuracy}
%\end{figure*}

%
%
%\begin{figure*} [ht]
%	\centering
%	\includegraphics[scale=0.45]{Figure/results/4nodes/node_loss_1.png}
%	\caption{Testing loss for different nodes in the learning process for SLHITL.}
%	\label{Fig_4nodes_loss}
%\end{figure*}

\subsubsection{Learning on eight nodes} 
\begin{figure} [ht]
 	\centering
	\includegraphics[width=.9\linewidth]{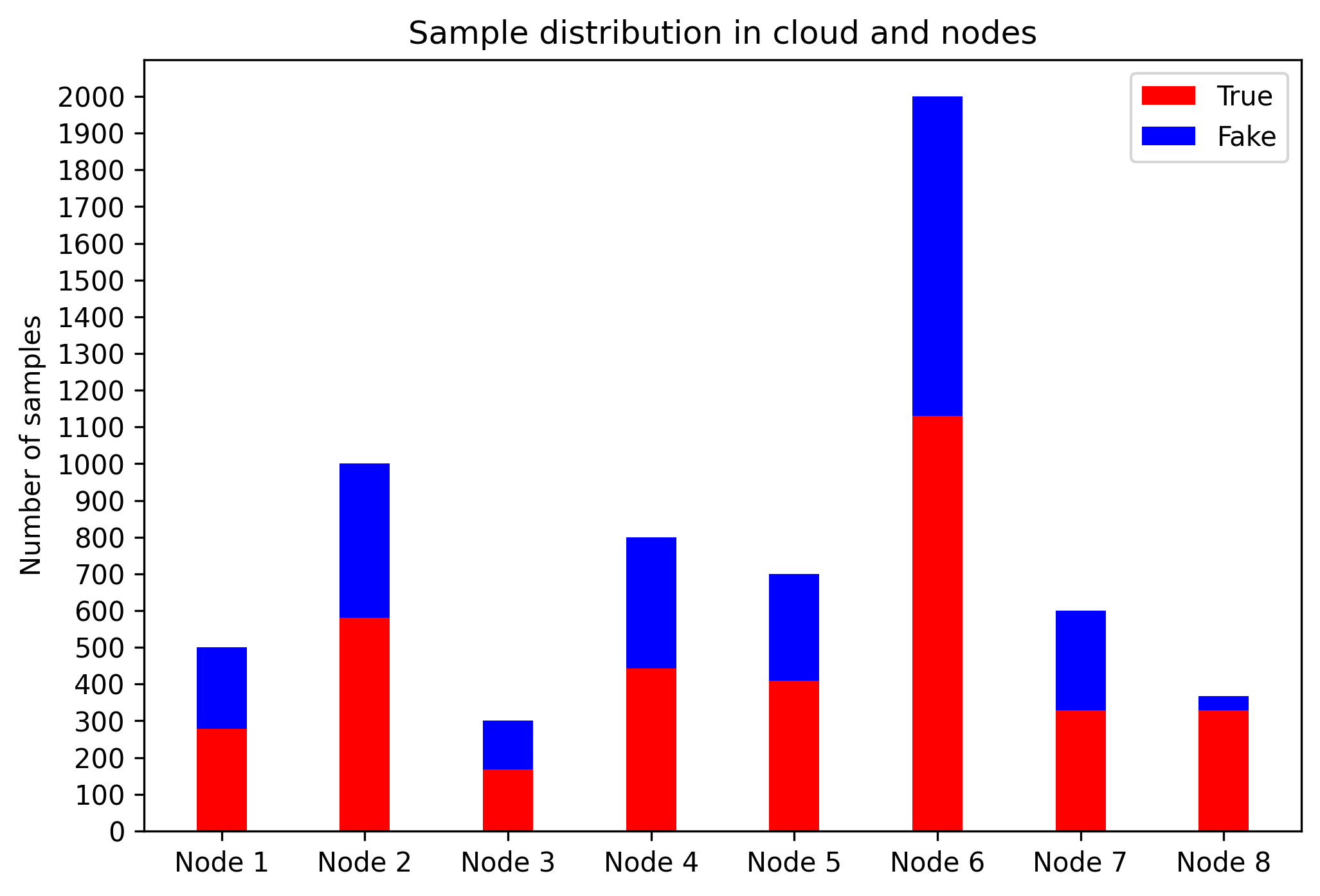}
	\caption{Sample distribution on 8 nodes for fake news detection. }
	\label{Fig_8_nodes_data}
\end{figure}

We also examine if HBSL can perform better than SL with the condition of learning on 8 nodes. Similarly, we distribute small number of samples for learning in different nodes as well, which is shown in Figure~\ref{Fig_8_nodes_data}. In addition to more data sets involved in this experiment, the size differences of samples for these data sets are larger compared to the case of learning on 4 nodes. For instance, the size difference between node 6 and node 3 is $1, 700$ ($2,000 - 300$) while that for the case of 4 node is $700$ ($1,000 - 300$) between node 2 and node 3. Moreover, for node 8, the dataset is not balanced. These changes will introduce new challenges like increasing converge time for fake news detection.

\begin{table*} [ht]
	\caption{  Accuracy (\%) comparison on fake news detection for learning with 8 nodes.  } 
        \begin{center}
                \begin{tabular}{|l|ccccccccc|}
                    \hline \textbf{SL} & Node 1 & Node 2 & Node 3 & Node 4 & Node 5 & Node 6 & Node 7 & Node 8 & \textbf{Average}\\ \hline
                    	   Run 1		&  87.99\%	&  84.39\%	&  93.33\%	&  85.50\%	&  89.14\%	&  89.80\%	&  88.66\%	&  80.00\%	&  87.35\%\\
                            Run 2		&  87.19\%	&  89.60\%	&  85.33\%	&  87.99\%	&  89.14\%	&  87.80\%	&  88.66\%	&  95.99\%	&  88.96\%\\
                            Run 3		&  81.59\%	&  88.40\%	&  90.66\%	&  89.99\%	&  87.42\%	&  87.99\%	&  93.33\%	&  92.00\%	&  88.92\%\\
                            Run 4		&  88.80\%	&  88.80\%	&  85.33\%	&  88.99\%	&  89.14\%	&  87.59\%	&  93.99\%	&  80.00\%	&  87.83\%\\
                            Run 5		&  89.60\%	&  90.79\%	&  90.66\%	&  89.99\%	&  86.28\%	&  87.59\%	&  88.66\%	&  83.99\%	&  88.45\%\\ \hline
                            \textbf{Average}		&  87.03\%	&  88.39\%	&  89.06\%	&  88.49\%	&  88.22\%	&  88.15\%	&  90.66\%	&  86.39\%	&  88.30\%\\  
                     \hline
                     \hline \textbf{HBSL} & Node 1 & Node 2 & Node 3 & Node 4 & Node 5 & Node 6 & Node 7 & Node 8 & \textbf{Average}\\ \hline
                    	   Run 1		&  89.60\%	&  90.79\%	&  90.66\%	&  89.99\%	&  86.28\%	&  87.59\%	&  88.66\%	&  83.99\%	&  88.45\%\\
                            Run 2		&  89.60\%	&  90.39\%	&  87.99\%	&  88.99\%	&  89.14\%	&  90.20\%	&  89.99\%	&  95.99\%	&  90.29\%\\
                            Run 3		&  85.60\%	&  88.80\%	&  92.00\%	&  91.50\%	&  91.42\%	&  87.59\%	&  95.99\%	&  95.99\%	&  91.11\%\\
                            Run 4		&  91.29\%	&  88.80\%	&  83.99\%	&  89.99\%	&  89.71\%	&  90.60\%	&  93.33\%	&  80.00\%	&  88.45\%\\
                            Run 5		&  93.59\%	&  91.60\%	&  92.00\%	&  91.50\%	&  90.28\%	&  89.60\%	&  89.33\%	&  87.99\%	&  90.73\%\\ \hline
                            \textbf{ Average}		&  89.93\%	&  90.07\%	&  89.32\%	&  90.39\%	&  89.36\%	&  89.11\%	&  91.46\%	&  88.79\%	&  89.80\%\\ 

                     \hline
                \end{tabular}
       \end{center}
         \label{tab_detail_sl_8nodes}
\end{table*}

 \begin{figure*} [ht]
	\centering
	\includegraphics[scale=0.22]{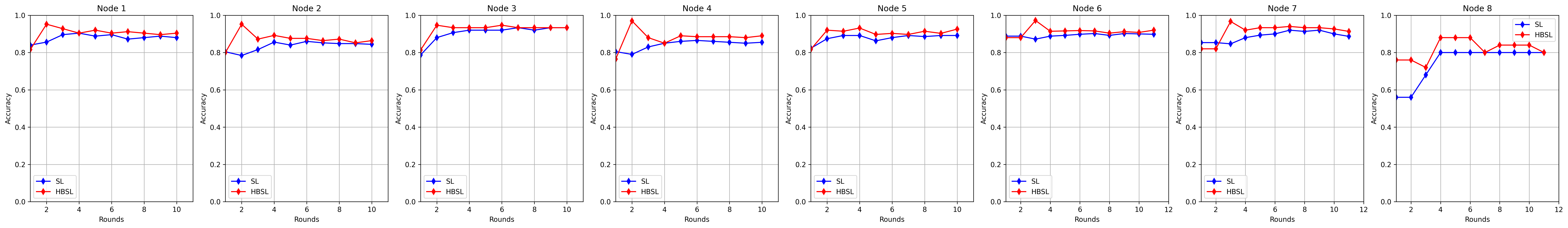} \\(a) Run 1\\
	\includegraphics[scale=0.22]{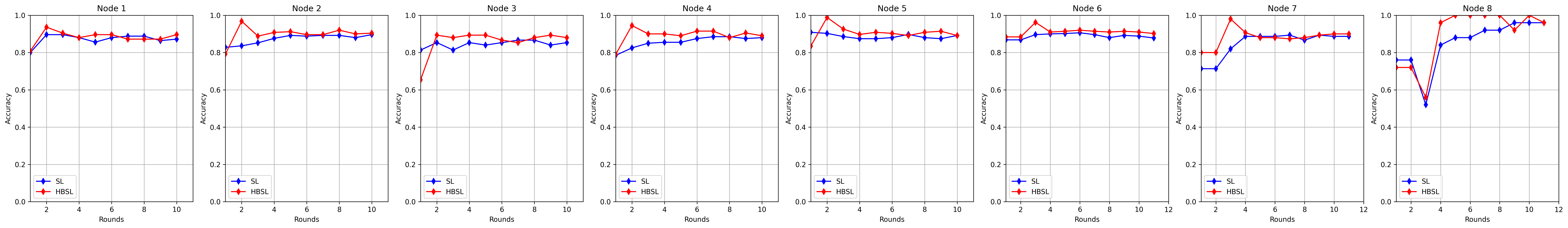} \\(b) Run 2 \\
	\includegraphics[scale=0.22]{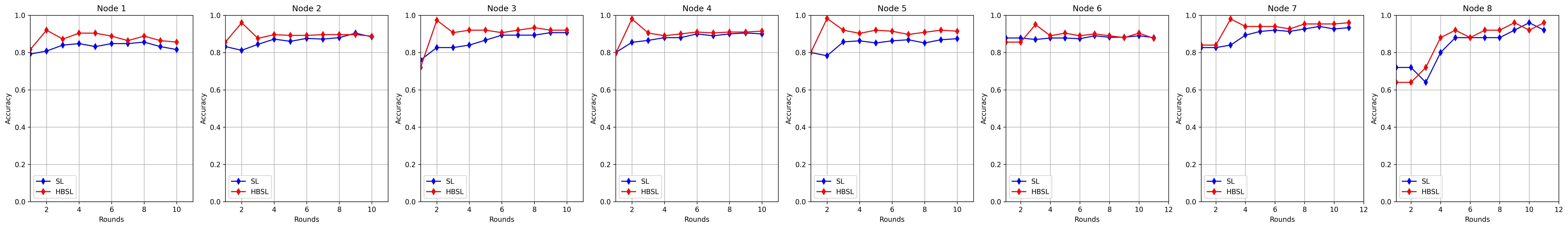} \\(c) Run 3\\ 
	\includegraphics[scale=0.22]{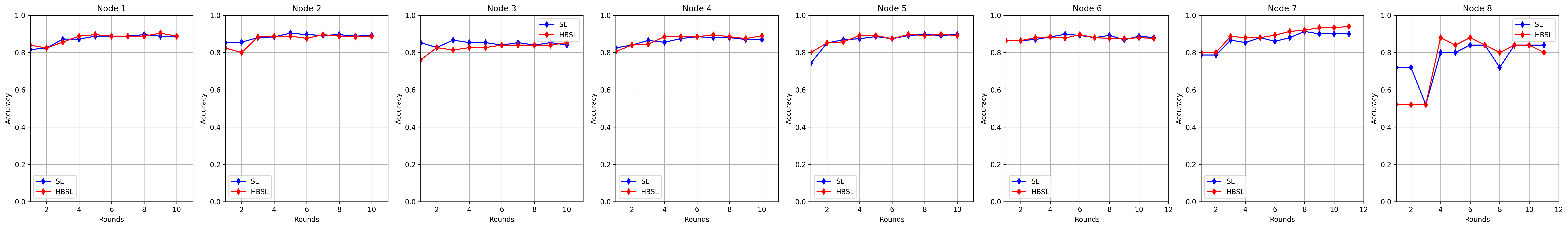} \\(d) Run 4\\
	\includegraphics[scale= 0.22]{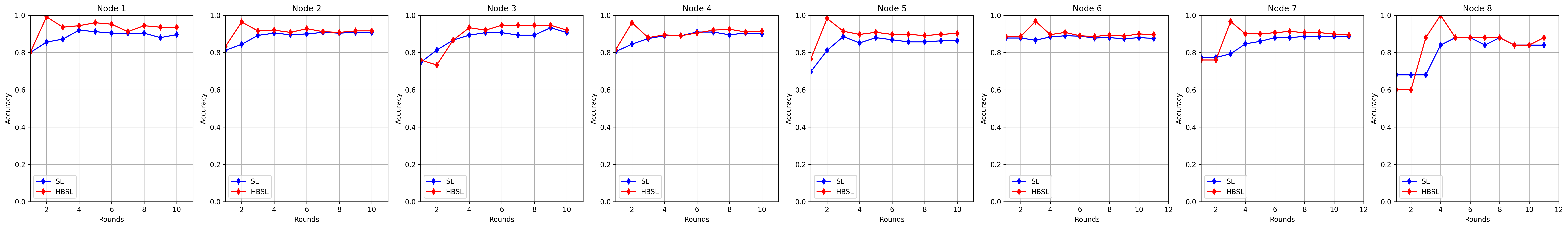} \\(e) Run 5\\
	\caption{Inference performance for 8 nodes in the learning process for HBSL in 5 runs.  Accuracy refers to testing accuracy.}
	\label{Fig_8nodes_accuracy}
\end{figure*}

Table~\ref{tab_detail_sl_8nodes} presents the comparison of the testing accuracy for the case with 8 nodes. Similar observation is obtained that HBSL outperforms SL on all nodes. However, the improvement is up to about 2\%, which is less that 5\% for the case of learning on 4 nodes. In other words, it seems that HBSL will be more suitable to learn on a local net with less nodes. Moreover, it enhances the detection performance for all runs in terms of average accuracy for different runs. Figure~\ref{Fig_8nodes_accuracy}  shows the testing accuracy for 8 nodes in the learning process for 5 runs. It is observed that the curves of test accuracy is various since the sizes of data for fine-tuning models will be various in the loop of learning and inference. In addition, although some nodes performed worse in the learning of HBSL, they can still converge to higher or similar performance of SL. Finally, it uses more learning rounds to achieve model convergence as it involves more nodes to update models, which reduces the convergence speed.

%\begin{table*}
%	\caption{  Accuracy (\%) comparison on fake news detection for learning with 8 nodes.  } 
%        \begin{center}
%                \begin{tabular}{|l|ccccccccc|}
%                    \hline \textbf{Model} & Node 1 & Node 2 & Node 3 & Node 4 & Node 5 & Node 6 & Node 7 & Node 8 & \textbf{Mean}\\ \hline
%                    	   
%                  Initialization	& 84.03\%		& 84.13\%		& 84.79\%		& 83.17\%		& 83.26\%		& 83.94\%		& 84.09\%		& 83.59\%		&  83.79\%\\  
%                     \hline
%                     SL		& 58.39\%		& 58.43\%		& 63.79\%		& 59.42\%		& 58.69\%		& 53.69\%		& 58.47\%		& 62.19\%		&  59.13\%\\  
%                     \hline             	   
%                  SLHITL   		& \textbf{99.60\%}	& \textbf{99.61\%}		& \textbf{97.93\%}	& \textbf{99.82\%}	& \textbf{99.79\%}	& \textbf{99.31\%}		& \textbf{99.71\%}	& \textbf{95.39\%}	&  \textbf{98.93\%}\\ 
%                     \hline
%                \end{tabular}
%       \end{center}
%         \label{tab_detail_sl_8nodes}
%\end{table*}
%

%\begin{figure*} [ht]
%	\centering
%	\includegraphics[scale=0.45]{Figure/results/8nodes/node_accuracy_1.png}
%	\caption{Testing accuracy for different nodes in the learning process for SLHITL for the case with 8 nodes.}
%	\label{Fig_8nodes_accuracy}
%\end{figure*}
%

%\begin{figure*} [ht]
%	\centering
%	\includegraphics[scale=0.45]{Figure/results/8nodes/node_loss_1.png}
%	\caption{Testing loss for different nodes in the learning process for SLHITL for the case with 8 nodes.}
%	\label{Fig_8nodes_loss}
%\end{figure*}

\section{Related Work}
\label{sec3}
% related work

Fake news detection attracted lots of attentions to effectively preventing the dissemination of fake news. Traditional fake news detection based on machine learning is classified into three categories, namely, content feature based~\cite{long2017fake, ruchansky2017csi, oshikawa2018survey, yang2019fake}, propagation feature based~\cite{shu2020hierarchical}, and context feature based~\cite{shu2019beyond}. Recently, more dimensions have been exploited. The first dimension is to combine these features to enhance detection performance. Moti~\textit{et al.} represented news as a graph containing these three features and utilized graph convolutional networks (GCN) to classify the graph into fake or true~\cite{monti2019fake}. In addition, it is resilience to adversarial attacks since generating adversarial samples on three features is extremely challenging. Wang~\textit{et al.} used multimodal features to detect fake news by developing event adversarial neural networks. It consists of three components: a multimodal feature extractor, a fake news detector, and an event classifier. The feature extractor recognized multimodal features on text and pictures related to news as input to the fake news detector and the event classifier. Then the fake news detector will beat the event classifier in the adversarial learning process, which is extract the multimodal features, that are more useful to fake news detection, through the learning process. 

The second dimension is to fight against adversarial ``fake news" generated by AI techniques. For instance, Fung~\textit{et al.} proposed \textit{InfoSurgeon} to detect adversarial ``fake news" that is represented as a fine-grained knowledge graph of news with external knowledge bases~\cite{fung2021infosurgeon}. It utilized graph neural networks to determine if the knowledge graph of news is true or fake. Shu~\textit{et al.} developed \textit{$FactGen$} to generate high-quality news by leveraging external facts to enrich the news and improve the consistency between input and output. Moreover, they proposed \textit{$FactGen_{def}$} to detect these synthetic fake news with high performance~\cite{shu2021fact}. 

The third dimension is to reduce efforts of labeling big data for training detection models. For example, Wang~\textit{et al.} proposed WeFEND, a reinforcement learning method, that is to leverage user's reports as weak supervision to enlarge training sets for fake news detection~\cite{wang2020weak}. It is composed of three components including annotator, reinforce selector, and detector. The annotator labeled data with weak labels and the reinforce selector chooses weakly labeled data to extend training sets for fine-tuning the detector. 

The forth dimension is to detect fake news from the perspective of psychology. For example, Karami~\textit{et al.} exploited fake news detection through combining psychology and data science. They extracted five features about motivation of spreading fake news that present user's psychology, namely, uncertainty, emotions, lack of control, relationship enhancement, and rank, where the first three features are extracted with Linguistic Inquiry and Word Count (LIWC)~\cite{tausczik2010psychological}, and the other two are obtained based on user behaviors on social media like retweeting and following. These features are significantly different between users who spread fake news and users who spread true news. Moreover, these features can be combined with content features to improve detection performance~\cite{karami2021profiling}.

Although these dimensions have broadened and enhanced fake new detection, few of work has exploited how to introduce human feedback into decentralized fake news detection to implement privacy preserving for users. This paper combined human-in-the-loop techniques with swarm learning to exploit decentralized  fake news detection.
 
%Cheng~\textit{et al.} identified confounder-variables to figure out which user attributes cause users to spread fake news by unbiased  estimation of fake news sharing behavior~\cite{cheng2021causal}. 

%\section{Discussion}
%\label{sec6}
%\input{Discussion}

\section{Conclusion}
\label{sec7}
 
 In this paper, a novel decentralized model is proposed for detecting fake news through combining swarm learning and human-in-the-loop. The learning and inference forms a loop until meeting stop criteria by learning and predicting on local data.  Specifically, the human feedback is generated in this loop to extend training sets for improving detection performance. The proposed model is validated on a benchmark, LIAR dataset. Experimental results indicate that the proposed model could outperform swarm learning on fake news detection in a decentralized manner. In the future, we plan to extend this work by designing detection models according to node features.

%\section*{Acknowledgment}
%\label{acknowledgement}
%This research work is supported in part by the U.S. Office of the Under Secretary of Defense for Research and Engineering (OUSD(R\&E)) under agreement number FA8750-15-2-0119. The U.S. Government is authorized to reproduce and distribute reprints for governmental purposes notwithstanding any copyright notation thereon. The views and conclusions contained herein are those of the authors and should not be interpreted as necessarily representing the official policies or endorsements, either expressed or implied, of the Office of the Under Secretary of Defense for Research and Engineering (OUSD(R\&E)) or the U.S. Government.

\ifCLASSOPTIONcaptionsoff
  \newpage
\fi


\bibliographystyle{IEEEtran}
\bibliography{References}
\end{document}